\documentclass{svproc}
\usepackage{url}
\usepackage{xcolor}
\usepackage{amsmath}
\usepackage{amsfonts}
\usepackage{amssymb}

\usepackage{graphicx}
\usepackage{subcaption}
\usepackage{multirow, multicol}
\usepackage{hyperref}

\graphicspath{{images/}}

\newcommand{\changes}[1]{\textcolor{black}{#1}}

\begin{document}
\mainmatter             

\title{Geometric analysis of generic 3R robots, and necessary and sufficient conditions for a class of orthogonal robots to have four IKS}
\titlerunning{Geometric analysis of 3R positional robots} 
\author{Durgesh Haribhau Salunkhe\inst{1} \and Abhilash Nayak\inst{2}}
\authorrunning{Salunkhe D. H. and Nayak A.} 

\institute{
    \email{salunkhedurgesh@outlook.com}
\and
    \email{abhilash.un@outlook.com}
}

\maketitle           

\begin{abstract}
The kinematic analysis of a generic 3R robot has been investigated with multiple approaches in the past. The algebraic approaches have established concrete results but are unfortunately limited to special classes or architectural simplifications. Geometric approaches on the other hand have extended the analysis to generic robots while also providing an intuitive understanding of their kinematic properties. We use the best of both approaches to present the kinematic analysis of a generic 3R robot, using the inverse kinematic model inherited from a method based on conformal geometric algebra. The paper discusses a generic framework to study the conditions for a 3R robot to have four inverse kinematic solutions (IKS) and allows to study the distribution of IKS as seen in workspace. The necessary and sufficient conditions for a class of orthogonal robots are presented using the proposed approach. 
% In this article, we present the kinematic analysis of a generic 3R robot, using the inverse kinematic model based on conformal geometric algebra. 

\end{abstract}
\section{Introduction}
\label{sec:introduction}
\iffalse
    \begin{itemize}
        \item Motivation
        \begin{itemize}
            \item History on kinematic analysis of 3R positional robots, (Pieper, Wenger, Thomas)
            \item CGA for ikin of 3R robots (Selig, hugo, isiah)
            \item Why number of IKS is important to study (little bit of cuspidal robots)
            \item why geometric analysis for generic results over pure algebraic methods
        \end{itemize}
    
        \item Contribution
        \begin{itemize}
            \item geometric interpretation of IKM of generic 3R positional robots using CGA
            \item Relationship between D-H parameters and the nature of the maximum number of IKS
            \item Necessary and sufficient condition for simplified geometries
            \item sufficient condition for orthogonal robot with $d_2, d_3 \neq 0$
        \end{itemize}
        
    \end{itemize}
\fi
The kinematic analysis of generic positional 3R robots was first addressed by Pieper~\cite{pieper_kinematics_1968} in 1968. Since then, the 3R robots have been studied by multiple approaches including geometric interpretation of inverse kinematic model (IKM) as intersecting conics, computer algebra, and algebraic analysis~\cite{wenger_condition}. These approaches have brought many insights into the nature of generic and nongeneric 3R robots. The condition for orthogonal 3R robots (with $d_3=0$) to have four inverse kinematic solutions (IKS) was presented in~\cite{wenger_condition} while the condition for the IKM to split in two quadratic polynomials was recently presented in~\cite{thomas_ellipse_quadratic}. Geometric approach has shown the ability to extend the conditions on cuspidality or number of IKS to generic robots~\cite{ark22_salunkhe}.\\
Recently, methods using conformal geometric algebra to investigate robot kinematics have gained popularity. In \cite{nayak_2025}, the IKM is interpreted as the intersection between a rotating and a fixed circle. In this work, we present the kinematic analysis of generic 3R robots by parameterizing the rotating circle as a torus. The advantage of this approach is that it gives a more straightforward geometric intuition of the IKM allowing better understanding of the effect of robot parameters on the number of IKS. This paper shows how the intersection of a circle with a torus-geometric manifolds that naturally appear as a link between the robot's workspace and joint space enables us to understand the conditions for a generic 3R robot to have four IKS.
Furthermore, as the torus is parameterized by the last two joint axes, the singularities in joint space can be plotted on this torus to gain better insights on the distribution of IKS as seen in the workspace. As an application of the analysis, we present the necessary and sufficient conditions for a class of orthogonal robots ($d_2=0, d_3 \neq 0$) to have four IKS \changes{extending the previously known results for robots with conditions ($d_2 \neq 0, d_3 = 0$)~\cite{wenger_condition}}. Additionally, we provide the limits on $z$ to transition from having four IKS to two IKS \changes{which is not available in previous analyses}. 

\section{Kinematic analysis of generic 3R robots}
\label{section:generic3r}
In this section, we discuss the kinematic analysis of a generic 3R robot and present the geometric classification for binary (robots with two IKS) and quaternary (robots with four IKS) robots.

\begin{figure}[htb]
    \centering
    \includegraphics[width=0.7\textwidth]{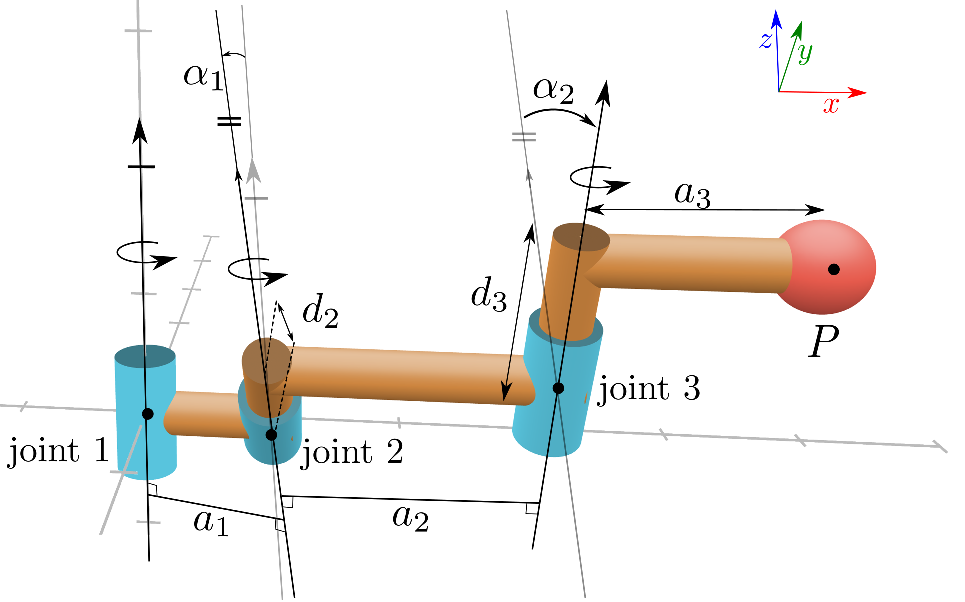}
    \caption{D-H parameters used to define the robot architecture}
    \label{fig:dh_params}
\end{figure}

\begin{figure}[tb]
    \centering
    \begin{subfigure}{0.48\textwidth}
        \centering
        \includegraphics[width=\textwidth]{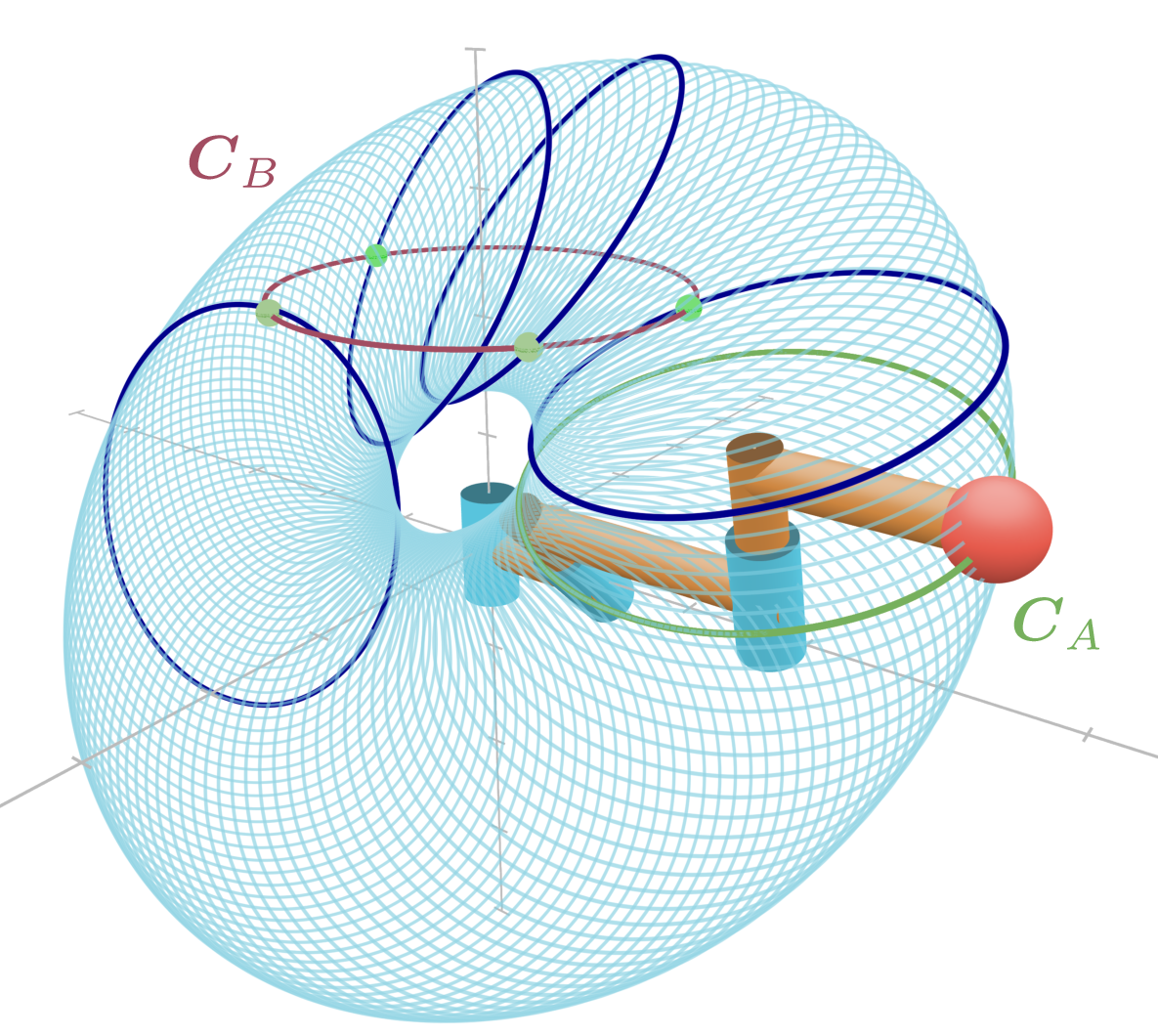}
    \caption{Schematic of IKM as intersection between torus traced by $C_A$, and circle $C_B$.}
    \label{fig:torus_circlesA}
    \end{subfigure}
    ~
    \begin{subfigure}{0.48\textwidth}
        \centering
        \includegraphics[width=\textwidth]{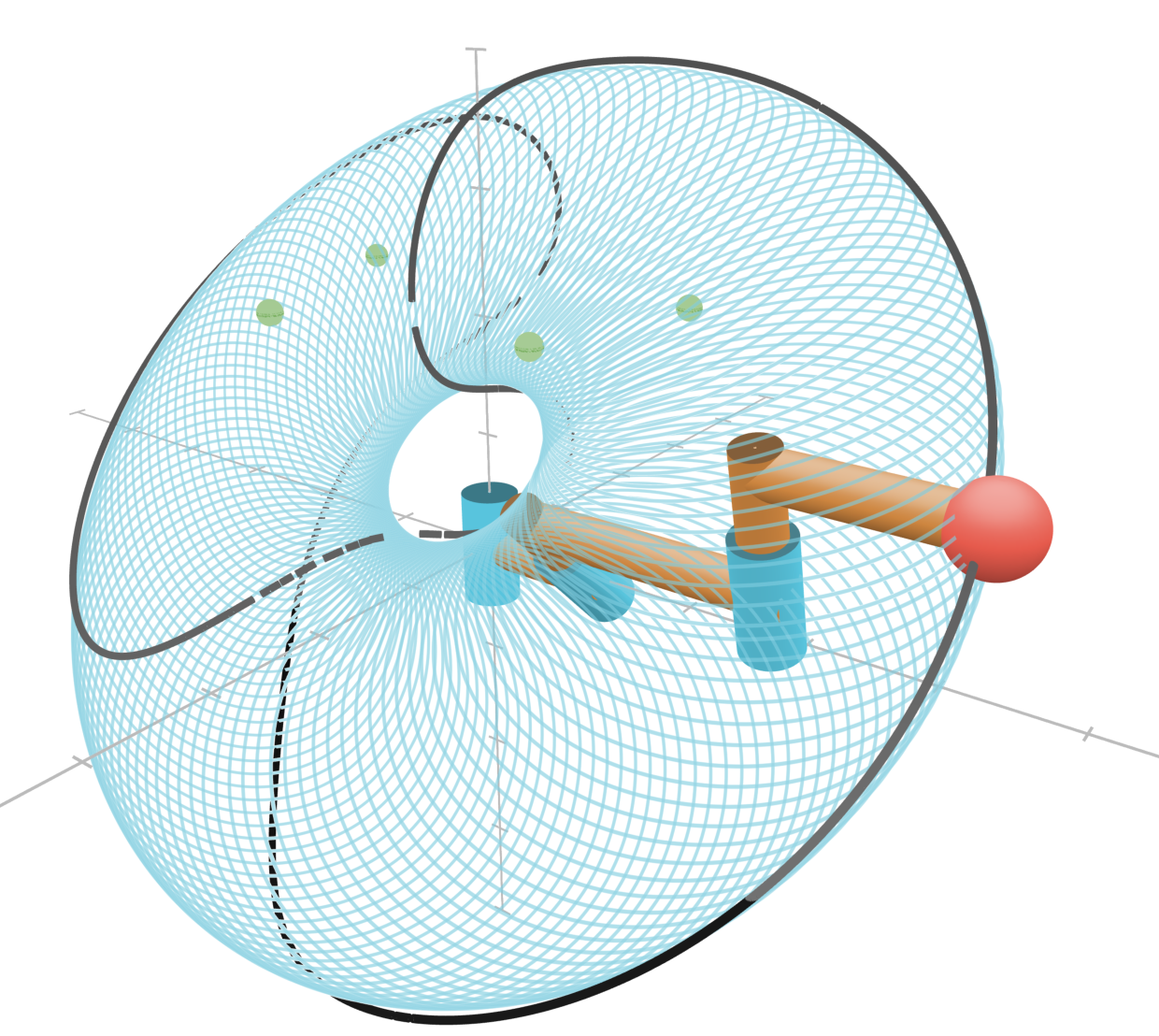}
    \caption{IKS distribution and singularities in jointspace visualized in workspace.}
    \label{fig:torus_circlesB}
    \end{subfigure}

    \caption{The illustration of the geometric interpretation of generic IKM and the visualization of distribution of IKS in workspace.}
    \label{fig:generic_ki_analysis}    
\end{figure}

\begin{figure}[tb]
    \centering
    \includegraphics[width=0.5\textwidth]{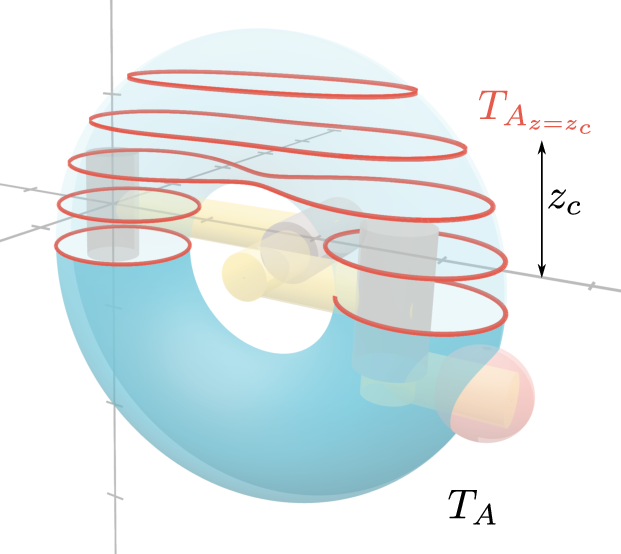}
    \caption{Torus seen as a combination of spiric sections along $z$-axis}
    \label{fig:torus_slices}
\end{figure}

A positional 3R robot is a kinematic chain with three rotating axes, and can be minimally represented by the Denavit-Hartenberg (D-H) parameters as shown in Fig.~\ref{fig:dh_params}. 
Following Selig's approach~\cite{selig}, we have shown in~\cite{nayak_2025} that the inverse kinematics of a generic positional 3R robot given an end-effector position can be reduced to finding the intersection of a fixed circle $C_B$ and a rotating circle $C_A$. To further analyze the number of IKS, it is useful to consider the torus $T_A$ traced by the rotating circle $C_A$ and find its intersection with the fixed circle $C_B$ as shown in Fig.~\ref{fig:torus_circlesA}. 
To look for the intersections of $T_A$ with $C_B$, we need only examine the cross-section of the torus that lies on the same plane as the circle, $C_B$. Let us identify this plane as $z=z_c$ given that $C_B$ is always parallel to the $xy$-plane. To determine the conditions on D-H parameters so that the 3R robot has 4 IKS, we should have four real intersections between $T_{A_{z=z_c}}$ and $C_B$. Figure \ref{fig:torus_slices} shows the toric sections at multiple $z=z_c$. It suffices to find one end-effector position, $P= (x, y, z - d_1)$ for which there are four intersections. 
Since the radius of $C_B$ depends on $P$, we look for points on the curve $T_{A_{z=z_c}}$ that are tangential to the circle $C_B$. We further check the conditions for the tangential circles to intersect the toric section in two other points. This means that we can always change the radius slightly to have two other intersections close to the tangential intersection leading to four total intersections. This way, we can cover all possible ways to get four intersections, thus obtaining the necessary and sufficient conditions for the robot to be quaternary.

\subsection{Orthogonal 3R robots}
\label{subsection:orthogonal3r}
\begin{figure}[htbp]
    \centering
    \includegraphics[width=0.85\textwidth]{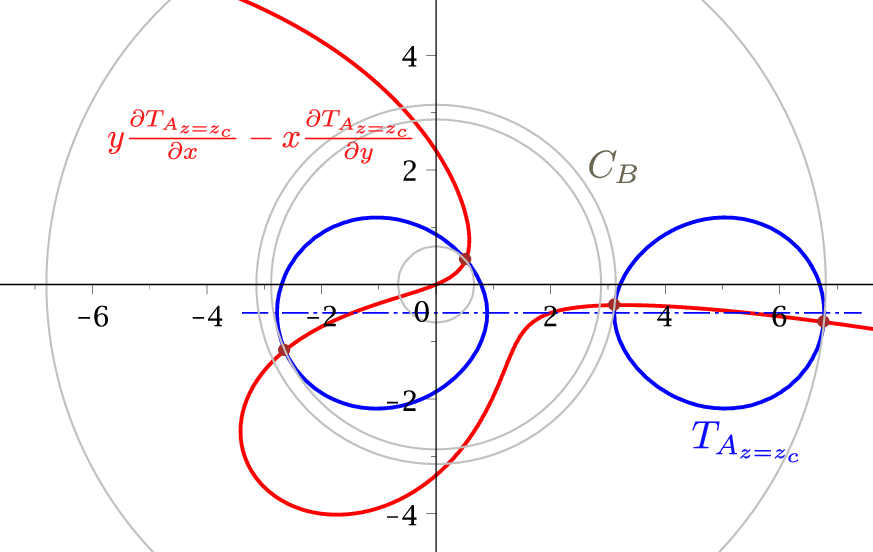}
    \caption{Tangential intersections of $C_B$ in gray with $T_{A_{z=z_c}}$ in blue and the slope equation in red.}
    \label{fig:tang_int}
\end{figure}

\begin{figure}[tb]
    \centering
    \includegraphics[width=0.9\textwidth]{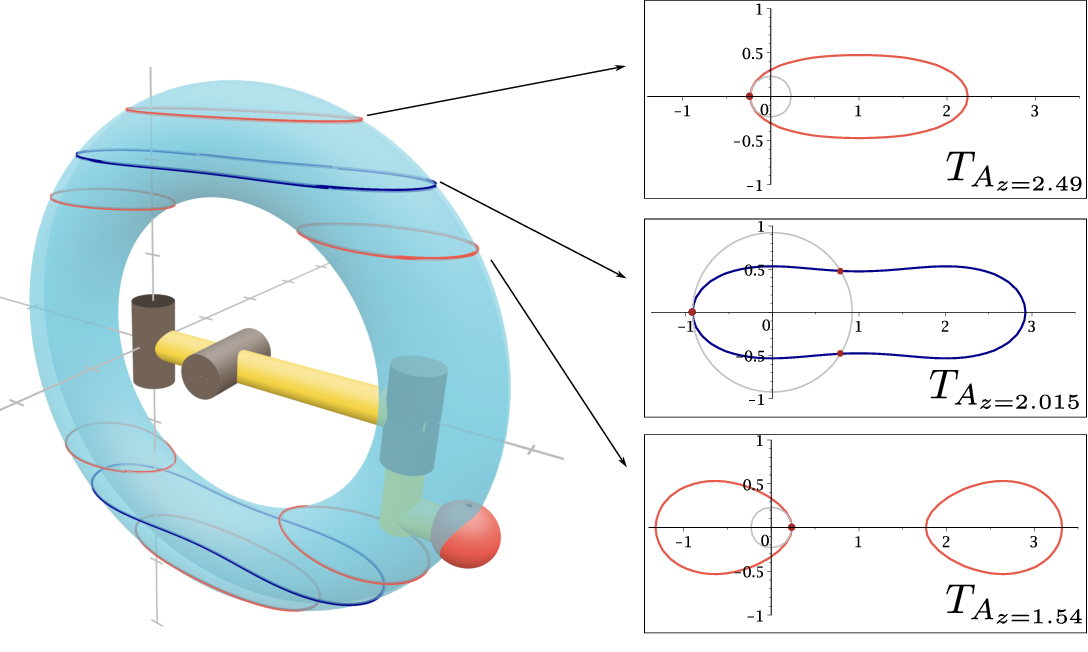}
    \caption{The limits of $z_c$ such that $T_A$ intersects $C_B$ at four distinct points.}
    \label{fig:torus_slices_transition}
\end{figure}

\begin{figure}[htbp]
    \centering
    \includegraphics[width=0.9\textwidth]{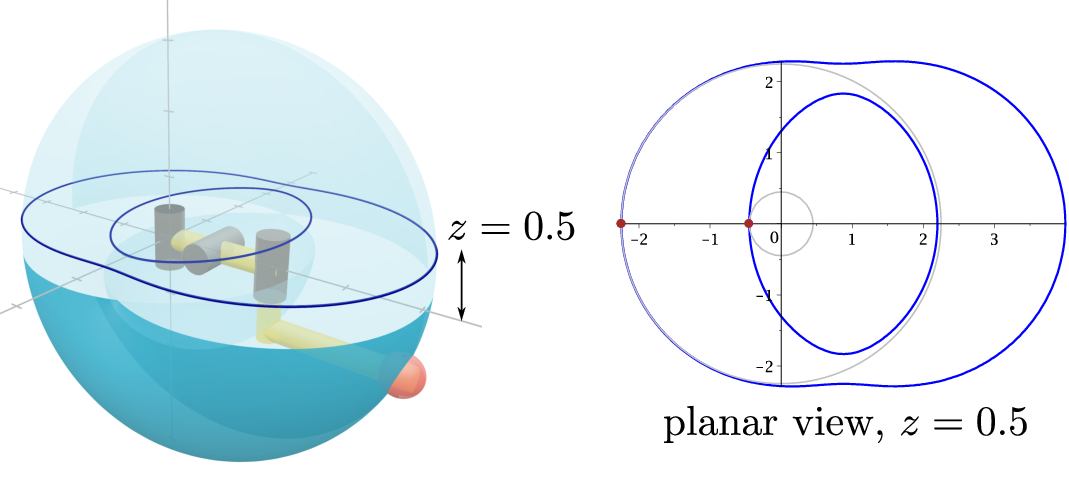}
    \caption{An example of a binary 3R robot with spindle torus and $a_1 = R$.}
    \label{fig:spindle_torus}
\end{figure}

In this section, we use the approach presented above for orthogonal robots, i.e., robots with a perpendicular joint arrangement ($\alpha_1 = \alpha_2 = \frac{\pi}{2}$) and present our analysis by taking $d_1 = 0$ without loss of generality. For an orthogonal 3R robot, the torus has a circular cross section and its axis is parallel to $y$-axis. Its center is $(a_1, -d_2, 0)$, minor radius $r = \frac{1}{2} \left( \sqrt{(a_2 + a_3)^2 + d_3^2} - \sqrt{(a_2 - a_3)^2 + d_3^2} \right)$ and major radius $R = \frac{1}{2} \left( \sqrt{(a_2 + a_3)^2 + d_3^2} + \sqrt{(a_2 - a_3)^2 + d_3^2} \right)$. Therefore, its implicit algebraic equation is:
\begin{equation} \label{eq:T_A}
    T_A : \left(  \left( x-a_{{1}} \right) ^{2}+{z}^{2}+{(y+d_2)}^{2}+{R}^{2}-{r}^{2}
 \right) ^{2}-4\,{R}^{2} \left(  \left( x-a_{{1}} \right) ^{2}+{z}^{2}
 \right) = 0
\end{equation}
The equation of $C_B$ on the plane $z=z_c$ is given by:
\begin{equation} \label{eq:C_B}
    C_B: x^2 + y^2 - r_B^2 = 0, \, z = z_c
\end{equation}
Equating the slopes of tangents at $(x, y)$ for $T_{A_{z=z_c}}$ and $C_B$ yields:
\begin{align}
    & -\frac{\partial C_B/ \partial x}{\partial C_B/ \partial y} = - \frac{2x}{2y} = -\frac{\partial T_{A_{z=z_c}}/ \partial x}{\partial T_{A_{z=z_c}}/ \partial y}  \Rightarrow y \frac{\partial T_{A_{z=z_c}}} {\partial x} - x \frac{\partial T_{A_{z=z_c}}} {\partial y} = 0 \label{eq:slope} \\
& \Rightarrow -
4\, \left( -x+a_{{1}} \right)  \left( -{R}^{2}-{r}^{2}+{x}^{2}-2\,xa_{
{1}}+{y}^{2}+2\,yd_{{2}}+{z_c}^{2}+{a_{{1}}}^{2}+{d_{{2}}}^{2} \right) y = 0 \nonumber
\end{align}
Fig.~\ref{fig:tang_int} shows the necessary tangential intersections for an example.
We would first like to determine radii of circles $C_B$ that are tangential to $T_{A_{z=z_c}}$. This is done by finding the Gr\"obner basis of the ideal of polynomials in~\eqref{eq:T_A}, \eqref{eq:C_B} and \eqref{eq:slope} at $z=z_c$ with the lexicographic ordering $x <_{lex} y <_{lex} < r_B$. It yields a 16 degree polynomial in $r_B$ with 1185 terms. As this system is complicated to analyse, we choose a special condition of $d_2 = 0$ to present a simpler analysis. For $d_2 = 0$, the torus is symmetric about the $x$-axis and the red curve in Fig.~\ref{fig:tang_int} splits into a circle and a line. It can be verified by substituting $d_2=0$ in~\eqref{eq:slope} that it factors and one of the factors is the line $y=0$, representing $x$-axis. The Gr\"obner basis can be recalculated with this substitution but it is straightforward to substitute $y=0$ in~\eqref{eq:T_A} and solve for $x$ which yields four solutions that represent radii $r_B$ of the outermost and innermost tangential circles:
\begin{equation}
    r_B = a_1 \pm \sqrt{(R \pm r)^2 - z_c^2}
\end{equation}
Now to determine if these circles intersect $T_{A_{z=z_c}}$ for values of $y \neq 0$, we substitute $r_B = a_1 \pm \sqrt{(R + r)^2 - z_c^2}$ in~\eqref{eq:C_B} and eliminate $x$ from $C_B$ and $T_{A_{z=z_c}}$ to obtain:
\begin{equation}
    \begin{aligned}
     & y^2 (( {R}^{4}-2\,{a_{{1}}}^{2}{R}^{2}+{a_{{1}}}^{4} ) {y}^{2}+
4\,{a_{{1}}}^{4}{R}^{2} \pm 4\,{a_{{1}}}^{4}Rr \pm 8\,{a_{{1}}}^{3}\sqrt {
  (R + r)^2-z_c^2}{R}^{2}  \\
  & \ \ +4\,{a_{{1}}}^{3
}\sqrt { (R + r)^2-z_c^2}Rr+4\,{a_{{1}}}
^{2}{R}^{4} \pm 12\,{R}^{3}r{a_{{1}}}^{2}+8\,{a_{{1}}}^{2}{R}^{2}{r}^{2}-4
\,{a_{{1}}}^{2}{R}^{2}{z_c}^{2} \\
  & \ \ +4\,a_{{1}}\sqrt {(R + r)^2-z_c^2}{R}^{3}r) = 0
\end{aligned} \label{eq:solve_for_y}
\end{equation}
Since we are interested in solutions other than $y=0$, solving for $y_s=y^2$ from the second factor and substituting $w^2 = (R + r)^2 - z_c^2$ to eliminate the square root, leads to:
\begin{equation}
    y_s = -4\,{\frac {a_{{1}}R \left( Rr \pm wa_{{1}}+{a_{{1}}}^{2} \right) 
 \left( \pm wR+a_{{1}}R + ra_{{1}} \right) }{ \left( a_{{1}}-R \right) ^{2}
 \left( a_{{1}}+R \right) ^{2}}}
\end{equation}
Assuming $a_1 \neq R$, $y_s$ must be positive for the existence of four solutions since $y = \sqrt{y_s}$. We have two possibilities:\\
(i) $r_B = a_1 + \sqrt{(R + r)^2 - z_c^2}$. In this case, the robot is quaternary when the following inequality holds:\\
$q_1(w) = (Rr + wa_1 + a_1^2)(wR + a_1R + a_1r) < 0$\\
Since $a_1>0$ and $R>0$, $q_1$ is always positive and the above condition is never satisfied. So, this case is discarded.\\
(ii) $r_B = a_1 - \sqrt{(R + r)^2 - z_c^2}$. In this case, the following inequality must be satisfied:\\
$q_2(w) = (Rr - wa_1 + a_1^2)(-wR + a_1R + a_1r) < 0$\\
Furthermore, by solving $q_1(w)=q_2(w)=0$ for $w$ and substituting the solutions in $z_c^2 = (R + r)^2 - w^2$, the critical values of $z_c^2$ are obtained as $$ z_1^2 = -{\frac { \left( {\it a_1}-r \right)  \left( {\it a_1}+r \right) 
 \left( {\it a_1}-R \right)  \left( {\it a_1}+R \right) }{{{\it a_1}}^{2}
}}, z_{2a}^2 = -{\frac { \left( R + r \right) ^{2} \left( {\it a_1}-R \right)  \left( {
\it a_1}+R \right) }{{R}^{2}}}$$
In addition, $0 \leq z_c < R + r$, and hence $0 \leq w < R + r$. Substituting these critical values of $w$ in $q_2(w)$ yields: $$q_2(0) = a_1(R+r)(a_1^2+Rr), \ q_2(R+r) = (R+r)(a_1 -   R)^2(a_1-r)$$
We need to check if $q_2(0)$ and $q_2(R+r)$ are negative as there could be values of $w$ within a tighter range than the one dictated by $q(w)=0$ that lead to a tighter range of $z$ for the existence of 4 IKS.\\
Following a similar procedure for $r_B = a_1 \pm \sqrt{(R - r)^2 - z_c^2}$, let us consider\\
(iii) $r_B = a_1 + \sqrt{(R - r)^2 - z_c^2}$. Here, the following inequality must be satisfied:\\
$q_3(w) = (-Rr + wa_1 + a_1^2)(+ wR - a_1R + a_1r) < 0$\\
Furthermore, substituting critical values of $0<w<|R-r|$ in $q$ yields:
\begin{align*}
   q_3(0) &= a_1(R-r)(a_1^2-Rr), \\ 
   q_3(R-r) &= (R-r)(a_1+R)^2(a_1-r) \mbox{ when } R > r \\
   q_3(r-R) &= (R-r)(a_1-R)^2(a_1+r) \mbox { when } r > R
\end{align*}
(iv) $r_B = a_1 - \sqrt{(R - r)^2 - z_c^2}$. Here, the following inequality must be satisfied:\\
$q_4(w) = (-Rr - wa_1 + a_1^2)(- wR - a_1R + a_1r) < 0$\\
Furthermore, substituting critical values of $0<w<|R-r|$ in $q$ yields:
\begin{align*}
   q_4(0) &= a_1(R-r)(a_1^2-Rr), \\ 
   q_4(R-r) &= (R-r)(a_1-R)^2(a_1+r) \mbox{ when } R > r \\
   q_4(r-R) &= (R-r)(a_1+R)^2(a_1-r) \mbox { when } r > R
\end{align*}
In both the above cases (iii) and (iv), by solving $q(w)=0$ for $w$ and substituting the solutions in $z_c^2 = (R - r)^2 - w^2$, the critical values of $z_c^2$ are  $$ z_1^2 = -{\frac { \left( {\it a_1}-r \right)  \left( {\it a_1}+r \right) 
 \left( {\it a_1}-R \right)  \left( {\it a_1}+R \right) }{{{\it a_1}}^{2}
}}, z_{2b}^2 = -{\frac { \left( R - r \right) ^{2} \left( {\it a_1}-R \right)  \left( {
\it a_1}+R \right) }{{R}^{2}}}$$
\begin{table}[htbp]
\centering
\resizebox{\textwidth}{!}{
\begin{tabular}{|c|c|c|c|c|c|c|c|c|c|c|c|}
\hline
Cases & $z_1^2$ & $z_{2a}^2$ & $z_{2b}^2$ & $q_2(0)$ & $q_2(R+r)$ & $q_3(0)$ & $q_3(R-r)$ & $q_3(r-R)$ &  $q_3(0)$ & $q_4(R-r)$ & $q_4(r-R)$\\
\hline
$a_1 < R < r$  & $-$ & $+$ & $+$ & $+$ & $-$ & $+$ & & $-$ & $+$ & & $+$\\
$a_1 < r < R$  & $-$ & $+$ & $+$ & $+$ & $-$ & $-$ & $-$ & & $-$ & $+$ & \\
$R < a_1 < r$  & $+$ & $-$ & $-$ & $+$ & $-$ & $-s$ & & $-$ & $-s$ & & $+$ \\
$R < r < a_1$  & $-$ & $-$ & $-$ & $+$ & $+$ & $-$ & & $-$ & $-$ & & $-$ \\
$r < a_1 < R$  & $+$ & $+$ & $+$ & $+$ & $+$ & $s$ &  $+$ & & $s$ & $+$ & \\
$r < R < a_1$  & $-$ & $-$ & $-$ & $+$ & $+$ & $+$ & $+$ & & $+$ & $+$ & \\
\hline
\end{tabular}
}
\caption{Signs of $z_i^2$ and $q_i(w)$ for 3R robot cases, where $s=\mbox{sign}(a_1^2$-$Rr)$}
    \label{tab:qsigns}
\end{table}
Let us consider one of the cases $a_1 < R < r$, listed in Table~\ref{tab:qsigns} to determine the conditions for the existence of 4 IKS: From Table~\ref{tab:qsigns}, $z_1^2<0$, $z_{2a}^2>0$ and $z_{2b}^2>0$. Hence, we expect the range of $z_c$ to be $\min(z_{2a}, z_{2b}) < z_c < \max(z_{2a}, z_{2b})$. From the expressions for $z_{2a}$ and $z_{2b}$, it is clear that $z_{2a} > z_{2b}$. Hence a tighter range would be $z_{2b} < z_c < z_{2a}$. We can further tighten the range by checking the signs of $q_i(w), i=\{2, 3, 4\}$ at $w=\{0, R+r, r-R\}$. Table~\ref{tab:qsigns} shows that $q_2(R+r)<0$ and $q_3(R-r)<0$, both of which correspond to the critical value of $z_c=0$. Considering that we only work on the positive $z=z_c$ plane, the final range is $0 \leq z_c < z_{2a}$. Note that the results should mirror about the $z=0$ plane, and the actual range should indeed be $-z_{2a} < z_c < z_{2a}$.  The necessary and sufficient conditions for the remaining cases can be derived in a similar fashion and they are all listed in Table~\ref{tab:all_conditions}.
An example of the case $r < a_1 < R$ is illustrated in Fig.~\ref{fig:torus_slices_transition} that shows the limits of $z_c$ and the transition from 2 IKS region to 4 IKS region.
%\vspace{-5mm}
%\vspace{-10mm}
~\\It is noteworthy that the case $r < R < a_1$ cannot lead to a quaternary robot since all $z^2$ values are negative and $q(w)$ values are positive. This condition is an improvement of the result from~\cite[Section 5]{wenger_condition} as it also holds for robots with $d_3 \neq 0$.
\begin{table}[tb]
    \centering
    \begin{tabular}{|c|c|}
\hline
  Cases & Conditions for 3R orthogonal robots to have 4 IKS  \\ % Header row
  \hline
  $a_1 < R < r$ & $0 \leq z_c < z_{2a}$\\
  $a_1 < r < R$ & $0 \leq z_c < \max(z_{2a}, R-r)$ \\
 $R < a_1 < r$ & $0 \leq z_c < z_1$ when $a_1^2 - Rr <0$, \\ 
 & $0 \leq z_c < \max(z_1, r-R)$ when $a_1^2 - Rr >0$  \\
$R < r < a_1$ & $0 \leq z_c < r-R$ \\
$r < a_1 < R$ & $\min(z_1, z_{2b}) < z_c < \max(z_1, z_{2a})$ when $a_1^2 - Rr < 0$, \\
              & $\min(z_1, z_{2b}, R-r) < z_c < \max(z_1, z_{2a}, R-r)$ when $a_1^2 - Rr > 0$ \\
$r < R < a_1, \ a_1=R$ & Never \\
  \hline
    \end{tabular}
    \caption{Necessary and sufficient conditions for a 3R orthogonal robot to be quaternary when $d_2=0$.}
    \label{tab:all_conditions}
\end{table}
%\vspace{-10mm}
For the case $a_1=R$, it is easy to check that the robot is never quaternary since in~\eqref{eq:solve_for_y}, the factor other than $y^2$ is independent of $y$. This condition also leads to the degeneracy condition of the robot to have infinite IKS. Interestingly, $a_1 = R$ is the only condition when a spindle torus does not lead to a quaternary robot and an example of such a robot is given in Fig.~\ref{fig:spindle_torus}.
%\vspace{-5mm}

% \vspace{-10mm}
\vspace{5mm}
\section{Conclusions}
This paper presents a geometric approach leading to a simplified algebraic analysis to study the conditions for a generic 3R robot with four IKS. This work already solves for orthogonal robots with $d_3 \neq 0$ and the study of conditions for generic 3R robots is ongoing.
%
% ---- Bibliography ----
%
\bibliographystyle{unsrt}
\bibliography{ref.bib}

\begin{thebibliography}{1}

\bibitem{pieper_kinematics_1968}
Donald~Lee Pieper.
\newblock {\em The {Kinematics} of {Manipulators} {Under} {Computer}
  {Control}}.
\newblock PhD thesis, Stanford University, USA, 10 1968.

\bibitem{wenger_condition}
Philippe Wenger, Damien Chablat, and Maher Baili.
\newblock A {DH}-parameter based condition for {3R} orthogonal manipulators to
  have four distinct inverse kinematic solutions.
\newblock {\em Journal of Mechanical Design}, 127(1):150--155, 03 2005.

\bibitem{thomas_ellipse_quadratic}
Bertold Bongardt and Federico Thomas.
\newblock On ellipse intersections by means of distance geometry.
\newblock In Masafumi Okada, editor, {\em Advances in Mechanism and Machine
  Science}, pages 533--543, Cham, 2023. Springer Nature Switzerland.

\bibitem{ark22_salunkhe}
Durgesh Salunkhe, Jose Capco, Damien Chablat, and Philippe Wenger.
\newblock Geometry based analysis of {3R} serial robots.
\newblock In Oscar Altuzarra and Andr{\'e}s Kecskem{\'e}thy, editors, {\em
  Advances in Robot Kinematics 2022}, pages 65--72, Cham, 2022. Springer.

\bibitem{nayak_2025}
Abhilash Nayak and Durgesh~Haribhau Salunkhe.
\newblock Inverse kinematic model for generic 3r positional robots using
  conformal geometric algebra.
\newblock pages 1--8, 03 2025.
\newblock Available at: \url{https://salunkhedurgesh.com/ima_references.html}.

\bibitem{selig}
Jonathan Selig.
\newblock {\em Geometric fundamentals of robotics}.
\newblock Springer, New York, 2005.

\end{thebibliography}
\end{document}